\documentclass[sigconf]{acmart}

\usepackage{mathtools}
\usepackage{booktabs}
\usepackage{algorithm}
\usepackage{algpseudocode}
\usepackage{graphicx}
\usepackage{subcaption}
\usepackage{xspace}
\usepackage{tikz}
\usetikzlibrary{arrows.meta,positioning,shapes.geometric,fit,calc}

\theoremstyle{acmplain}
\newtheorem{lemma}{Lemma}[section]
\newtheorem{corollary}{Corollary}[section]
\newtheorem{proposition}{Proposition}[section]

\theoremstyle{acmdefinition}
\newtheorem{definition}{Definition}[section]

\theoremstyle{acmplain}

\newcommand{\KL}{D_{\mathrm{KL}}}

\newcommand{\AoInt}{\operatorname{AoInt}}
\newcommand{\Reg}{\operatorname{Reg}}
\newcommand{\airsync}{\textsc{Param-AoIntSync}\xspace}
\newcommand{\klairsync}{\textsc{KL-AoIntSync}\xspace}

\newtheorem{assumption}{Assumption}[section]

\AtBeginDocument{%
  }

\copyrightyear{2026}
\acmYear{2026}
\setcopyright{cc}
\setcctype{by}
\acmConference[MobiHoc '26]{The Twenty-seventh International Symposium on Theory, Algorithmic Foundations, and Protocol Design for Mobile Networks and Mobile Computing}{November 23--26, 2026}{Tokyo, Japan}
\acmBooktitle{The Twenty-seventh International Symposium on Theory, Algorithmic Foundations, and Protocol Design for Mobile Networks and Mobile Computing (MobiHoc '26), November 23--26, 2026, Tokyo, Japan}
\acmDOI{10.1145/3842721.3850123}
\acmISBN{979-8-4007-2975-1/2026/11}
\usepackage{etoolbox}
\begin{document}
\setlength{\abovedisplayskip}{2pt}
\setlength{\belowdisplayskip}{2pt}
\setlength{\skip\footins}{6pt}
\setlength{\footnotesep}{0pc}
\title{Fresh Enough to Decide: Age of Intelligence for Cost-Aware Model Synchronization in Dynamic IoT Systems}

% \usepackage{etoolbox}

\makeatletter

% Give the author row 20 mm more room on the RIGHT
%\patchcmd{\@mkauthors@iii}
%  {\hsize=\textwidth}
%  {\hsize=\dimexpr\textwidth+22mm\relax}
%  {}
%  {}

%\newcounter{myAuthorBox}
%\let\old@typeset@author@bx\@typeset@author@bx

%\def\@typeset@author@bx{%
%  \stepcounter{myAuthorBox}%
%  \ifnum\value{myAuthorBox}=1 \hspace{-23mm}\fi
%  \ifnum\value{myAuthorBox}=2 \hspace{-22.5mm}\fi
%  \ifnum\value{myAuthorBox}=3 \hspace{-14.5mm}\fi
%  \ifnum\value{myAuthorBox}=4 \hspace{-21mm}\fi
%  \ifnum\value{myAuthorBox}=5 \hspace{-26mm}\fi
%  \old@typeset@author@bx
%}

\makeatother
% \maketitle
\author{Hui Wan}
\affiliation{%
  \institution{University at Buffalo}
  \city{Buffalo}
  \state{NY}
  \country{USA}}
%\hspace{-16mm}}
\email{huiwan@buffalo.edu}
\orcid{0009-0008-8086-1332}

\author{Seyyedali Hosseinalipour}
\affiliation{%
  \institution{University at Buffalo}
  \city{Buffalo}
  \state{NY}
  \country{USA}
}
\email{alipour@buffalo.edu}
%\orcid{0009-0008-8086-1332}

%\author{Nicholas Mastronarde}
%\affiliation{%
%  \institution{University at Buffalo--SUNY}
%  \city{Buffalo}
%  \state{NY}
%  \country{USA}
%\email{nmastron@buffalo.edu}
%\orcid{0009-0008-8086-1332}

\author{Jacob Chakareski}
\affiliation{%
  \institution{New Jersey Institute of Technology}
  \city{Newark}
  \state{NJ}
  \country{USA}
}
\email{jacobcha@njit.edu}

\author{Ming Shi}
\correspondingauthor
\affiliation{%
  \institution{University at Buffalo}
  \city{Buffalo}
  \state{NY}
  \country{USA}
}
\email{mshi24@buffalo.edu}
\orcid{0000-0001-9941-9095}
\renewcommand{\shortauthors}{Wan and Shi}

\begin{abstract}
Resource-constrained IoT devices cache learned models for real-time decisions, but frequent updates incur substantial bandwidth, energy, and service overhead. Because decision-relevant model differences need not grow with elapsed time, we define \emph{Age of Intelligence} ($\AoInt$) as a model-version-relative, client-specific, and action utility distortion, and formulate synchronization as a  tradeoff between stale-decision regret and refresh cost. 
%Although the utility function is known for action selection, the synchronization trigger requires no realized outcomes, labels, or rewards. 
Furthermore, we develop Kullback-Leibler-aware AoInt Synchronization (\klairsync), which accumulates output-distribution changes using client-specific unlabeled probes, whereas Parameter-aware AoInt Synchronization (\airsync) uses a server-broadcast scalar parameter change. Then, we establish pathwise guarantees with fixed client distributions, action sets, and utilities. The numerical results show that calibrated \airsync reduces joint synchronization cost by $6.15\%$ relative to an Instant-Parameters policy while performing comparably to \klairsync.

\end{abstract}
%$\mathcal{O}(\sqrt{L_i c_i T V_T^\theta})$


% \includegraphics[width=\textwidth]{sampleteaser}


\maketitle

\section{Introduction}

The Internet of Things (IoT) increasingly relies on device-side machine-learning models for latency-sensitive prediction and control in applications such as industrial automation, transportation, health monitoring, spatial computing, and smart infrastructure~\cite{DOMINGUEZBOLANO2022100626,zhou2019edge,ChakareskiKY:20,NadimiABCH:25}. Executing inference near the data source can reduce response latency and raw-data exposure, and can sustain operation when cloud connectivity is intermittent. But, the environments observed by IoT devices are often dynamic~\cite{YANG2022105366}: sensing conditions, user behavior, equipment state, and task requirements may change over time. A model suitable when installed can consequently become misaligned with its present operating conditions. %under which a device must act.

Resource constraints make this problem especially challenging. Many devices can execute a compact inference model but cannot continuously retrain it, retain numerous versions, or repeatedly download large updates. A common architecture therefore updates models at an edge server and lets each device perform inference with a locally cached copy. This separation creates a downstream deployment decision distinct from model training: when should a device replace its cached model with the latest server version? Every refresh consumes bandwidth and energy, occupies memory, and may interrupt service while the new model is validated and installed. Conversely, postponing a consequential update can degrade the decisions made with stale intelligence. Communication-efficient distributed-training methods, including federated learning, adaptive aggregation, event-triggered SGD, and gossip-based learning~\cite{mcmahan2017fedavg,wang2019adaptivefl,george2020event,zhai2026event}, reduce the cost of producing a model but do not determine whether an already trained upstream model is worth deploying to a particular device.

A natural baseline is to refresh periodically or according to the elapsed time since the previous update. Age of Information (AoI) formalizes this chronological notion of freshness~\cite{kaul2012aoi,sun2017update,yates2021survey}. Yet the operational value of a model need not decay monotonically with time. During stable operation, an old model may remain prediction- or decision-equivalent to the latest version; after an abrupt change, even a recently installed model may become obsolete. This observation has motivated richer freshness metrics that incorporate correctness, estimation error, goals, or value~\cite{maatouk2020aoii,chen2020beyond,di2023goal,holm2023goal}. For learning services, Age of Model measures inference-performance degradation~\cite{chi2024aom}, while Age of Context measures the relevance of the samples supporting an edge-intelligence service~\cite{xu2023aoc}. These approaches establish that task impact matters, but they do not directly supply the low-rate signal needed for cost-aware model deployment.

In particular, a constrained device often holds only its cached model and cannot download and execute every candidate merely to assess its value. Such an assessment may require storing two model versions, evaluating representative inputs, and validating decisions before installation. Moreover, the labels or rewards needed to measure decision performance may be delayed, sparse, or unavailable online. Heterogeneous contexts, actions, utilities, and refresh costs also mean that a single server-side accuracy score may not characterize the value of an update for every device. An effective synchronization mechanism must detect consequential model evolution before deployment, operate without online supervision, accumulate gradual changes, and expose a tunable tradeoff between decision degradation and refresh cost.

We cast this synchronization task as a nonstationary decision problem with an explicit refresh cost, drawing on variation budgets, dynamic regret, and switching-cost learning~\cite{besbes2015nonstationary,cheung2019nonstationary,dekel2014switching,shi2025power,dd3b909d3d164cceb2e265c2284292b9}. We introduce \emph{Age of Intelligence} (AoInt), the expected utility loss caused by acting with a device's cached model rather than the current server model. Unlike elapsed time or raw parameter distance, AoInt directly measures the decision consequence of staleness and therefore distinguishes model motion from model changes that actually affect downstream actions.

\begin{figure}
    \centering
    \includegraphics[width=0.95\linewidth]{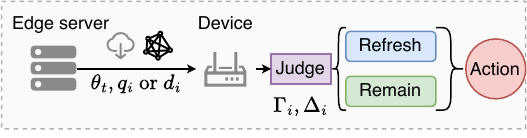} 
    \caption{Overview of the proposed AoInt-Aware synchronization framework.}
    \label{fig:bigpicture}
\end{figure}

Our contributions are threefold. First, we formulate cost-aware model synchronization using AoInt, a client-specific and action-utility distortion between the current server model and a device's cached model. We then derive a Kullback--Leibler (KL) output-space certificate satisfying $\AoInt_{i,t}=r_{i,t}(\theta_{i,t})\leq \kappa\widehat{C}_{i,t}^{\mathrm{KL}}$ \color{black}with high probability\color{black}, where $r_{i,t}$ denotes the instant regret of device $i$ under cached model parameter $\theta_{i,t}$ and $\widehat{C}_{i,t}^{\mathrm{KL}}$ denote the approximate KL divergence between cached model and server model. Second, we develop KL-aware Age-of-Intelligence Synchronization (\klairsync), which accumulates predictive-distribution changes evaluated on client-specific unlabeled probes, and its lightweight Parameter-aware Age-of-Intelligence Synchronization (\airsync), which uses a single scalar parameter-change signal broadcast by the server. For both methods, we establish pathwise bounds on stale-model regret, refresh count, and their joint cost, as well as sublinear regret when cumulative server-model variation is sublinear. 
%These guarantees hold without a stochastic model for the server-model sequence and accommodate both gradual drift and abrupt revisions, while each client's context distribution $\mu_i$, action set $\mathcal A_i$, and utility function $h_i$ remain fixed over the horizon but may differ across clients. 
Third, simulations with heterogeneous clients and server-model paths containing both gradual drift and abrupt jumps show that \airsync reduces joint cost by $6.15\%$ and regret by $9.22\%$ relative to a \color{black}Instant-Parameters \color{black}policy, while achieving performance comparable to \klairsync. The proposed framework and its key components are summarized in Fig.~\ref{fig:bigpicture}.

\section{Problem Formulation and Age of Intelligence}
\label{sec:model}

%\color{black}
We consider an IoT system with $N$ heterogeneous devices $\mathcal N=\{1,\ldots,N\}$ over a horizon $\mathcal T=\{1,\ldots,T\}$. \color{black}At each round $t\in\mathcal{T}$, the server maintains its latest model $\theta_t\in\Theta$, while device $i\in \mathcal{N}$ operates with a cached model $\theta_{i,t}\in\{\theta_\tau\}_{\tau \in\{1,\ldots,t\}}$. Before observing or acting, the device either retains its cached model or synchronizes with $\theta_t$ at cost $c_i$. Specifically, device $i$ makes a binary synchronization decision $z_{i,t}\in\{0,1\}$ and updates its model as $\theta_{i,t} = z_{i,t}\theta_t + (1-z_{i,t})\theta_\tau(\tau \in [1,t))$. Here, $z_{i,t}=0$ retains the cached model, whereas $z_{i,t}=1$ installs the current server model $\theta_t$ before the round-$t$ action and incurs a normalized refresh cost $c_i\in\mathbb R_{+}$. The total number of model refreshes is $M_i(T):=\sum_{t=1}^{T}z_{i,t}$. The synchronization sequence is generated by a policy $\pi:\Theta \times \Theta \mapsto\{0,1\}$, and $\Pi_c$ denotes the class of causal policies that use only information available up to round $t$.

Next, a latent context $s_{i,t}\in\mathcal{S}$ is drawn from $\mu_i$. Due to space limitations, we consider a fixed context distribution and leave extensions to time-varying distributions to future work. Conditional on this context, $\theta_{i,t}$ induces a predictive distribution $\mathbb P_{\theta_{i,t}}(y_{i,t}\mid s_{i,t})$ over the outcome space $\mathcal Y$. Based on the model output and its utility function $u_i:\mathcal{Y} \times \mathcal{A}\mapsto [0,1]$, device $i$ selects an action $a_{i,t}\in\mathcal A$. We impose the following assumption before introducing the performance metrics.

\begin{assumption}
\label{ass:system}
For each device $i\in\mathcal N$,
the distribution $\mu_i$, $\mathcal{S}$ with $|\mathcal{S}|=S<\infty$,
$\mathcal{A}$ with $|\mathcal A|=A <\infty$,
$\mathcal{Y}$ with $|\mathcal Y|=Y <\infty$, and
$c_i\in\mathbb R_{+}$ remain fixed over $\mathcal T$,
where $|\cdot|$ denotes cardinality.
\end{assumption}

\begin{definition}[Decision Utility, Regret, and AoInt]
\label{def:utility-regret}
For each device $i\in\mathcal N$, let
$h_i:\mathcal A_i\times\mathcal S\to[0,1]$
be a \color{black}known utility function\color{black}. The optimal expected cumulative reward under model $\theta$ is
$V_i^{\theta} = \sup_{a_{i,t}\in \mathcal{A}_i}\mathbb{E}_{s_{i,t}\sim\mu_i,y_{i,t}\sim\mathbb{P}_{\theta}(\cdot |s_{i,t})}[\sum\nolimits_{t\in \mathcal{T}}h_i(a_{i,t},y_{i,t})]$,
and the corresponding oracle value is
$V_i^{*} = \sup_{a_{i,t}\in \mathcal{A}_i}\mathbb{E}_{s_{i,t}\sim\mu_i}[\sum\nolimits_{t\in \mathcal{T}}h_i(a_{i,t},s_{i,t})]$.
\color{black}The regret of device $i$ can therefore be decomposed as
\begin{equation}
    \Reg_i = V^*_i - V^{\theta_{i,t}}_i = V^*_i - V^{\theta_{t}}_i + (V^{\theta_t}_i - V^{\theta_{i,t}}_i ).
\end{equation}
\color{black}This decomposition separates the oracle gap of the current server model from the additional gap associated with the cached model. Accordingly, we define AoInt as
\begin{equation}
    \AoInt_{i,t} = \sup\nolimits_{a\in \mathcal{A}_i} \mathbb{E}_{\theta_t}h_i(a,y_{i,t}) - \sup\nolimits_{a\in \mathcal{A}_i} \mathbb{E}_{\theta_{i,t}}h_i(a,y_{i,t}),
\end{equation}
where $\mathbb{E}_{\theta}$ denotes
$\mathbb{E}_{y_{i,t}\sim \mathbb{P}_\theta}$.
The cumulative AoInt bounds the corresponding value gap:
$\sum\nolimits_{t\in \mathcal{T}}\AoInt_{i,t} \geq V^{\theta_t}_i - V^{\theta_{i,t}}_i$.
\end{definition}

\color{black}
To focus on evaluating the synchronization process, we impose the following assumption on the current server model.

\begin{assumption}
\label{ass:optimal}
For each device $i$, the current server model $\theta_t$ is $\epsilon$-optimal that satisfies $V^*_i - V^{\theta_{t}}_i \leq \epsilon,\epsilon > 0$, so that
$\Reg_i \leq \epsilon + (V^{\theta_t}_i - V^{\theta_{i,t}}_i ) \leq \epsilon+ \sum_{t\in \mathcal{T}}\AoInt_{i,t}$.
\end{assumption}
\color{black}
AoInt measures a difference in expected action utility rather than an observed realized reward loss. It may be positive even when the maximizing action remains unchanged. Moreover, unlike chronological AoI, it need not increase monotonically between refreshes, because the server model may move toward or away from the cached version. The AoInt component measures deployment inconsistency relative to this reference, without implying that every newer model is necessarily better for the client's true environment. In the following, we write $r_{i,t} = \AoInt_{i,t}$.
\color{black}

\begin{definition}[Joint Cost]
\label{def:joint-objective}

The joint cost incurred by $\pi$ over the horizon combines stale-model regret and synchronization cost, where $r^\pi_{i,t}=r_{i,t}(\pi(\theta_t,\theta_{t-1})\theta_t+(1-\pi(\theta_t,\theta_{t-1}))\theta_{i,t})$:
\begin{equation}
    J_T^\pi:=
    \sum\nolimits_{i=1}^{N}\sum\nolimits_{t=1}^{T}
    \left[ r^\pi_{i,t}+c_i \pi(\theta_t,\theta_{t-1})\right].
    \label{eq:causal-objective}
\end{equation}
\end{definition}

\color{black}
\begin{definition}[Output-KL Certificate]
\label{def:kl-certificate}
For any models $\theta$ and $\theta'$, we define the KL certificate as 
\begin{equation}
C_i^{\mathrm{KL}}(\theta,\theta')
    :=
    \sqrt{\frac12\mathbb E_{s\sim\mu_i}
        \KL\!\left(
            p_\theta(\cdot\mid s)
            \middle\|
            p_{\theta'}(\cdot\mid s)
        \right)}.
        \end{equation}
Since the online context $s_{i,t}$ and its distribution $\mu_i$
are unknown, we estimate it by given probes
$\{s_i^{(m)}\}_{m=1}^{M_i}$, and estimate it by
\begin{equation}
    \widehat C_i^{\mathrm{KL}}(\theta,\theta')
    :=
    \sqrt{\frac{1}{2M_i}
        \sum\nolimits_{m=1}^{M_i}
        \KL\!\left(
            p_\theta(\cdot\mid s_i^{(m)})
            \middle\|
            p_{\theta'}(\cdot\mid s_i^{(m)})
        \right)}.
    \label{eq:kl-certificate}
\end{equation}
Henceforth, we write
$C^{\mathrm{KL}}_{i,t}
:=C_i^{\mathrm{KL}}(\theta_t,\theta_{i,t})$
and
$\widehat C^{\mathrm{KL}}_{i,t}
:=\widehat C_i^{\mathrm{KL}}(\theta_t,\theta_{i,t})$.
%The population certificate bounds AoInt, whereas its empirical
%counterpart requires a sampling-error correction to provide
%a population guarantee.
\end{definition}
\color{black}

Next, we bound the magnitude of AoInt using
the predictive-distribution discrepancy between the current
and cached models.

\begin{proposition}[Empirical KL Control of AoInt]
\label{prop:aoint-control}
Under Definitions~\ref{def:utility-regret}
and~\ref{def:kl-certificate}, fix $(\theta_t,\theta_{i,t})$
independently of $M_i$ i.i.d.\ probes from $\mu_i$.
Let $X(s):=\KL(p_{\theta_t}(\cdot\mid s)\|
p_{\theta_{i,t}}(\cdot\mid s))\in[0,B_i]$ almost surely
and $D:=\mathbb E_{\mu_i}X>0$.
For $\kappa>1$ and $\delta\in(0,1)$, if
$M_i\geq B_i^2\log(1/\delta)/
[2D^2(1-\kappa^{-2})^2]$, then, with probability
at least $1-\delta$,
$|\AoInt_{i,t}|=|r_{i,t}(\theta_{i,t})|
\leq C^{\mathrm{KL}}_{i,t}
<\kappa\widehat C^{\mathrm{KL}}_{i,t}$.
\end{proposition}

\begin{proof}
Bounded utilities, Pinsker's inequality,
$|\max f-\max g|\leq\max|f-g|$, and Jensen's
inequality give
$|\AoInt_{i,t}|\leq\mathbb E\sqrt{X/2}
\leq\sqrt{D/2}=C^{\mathrm{KL}}_{i,t}$.
Hoeffding's inequality gives
$\Pr(C^{\mathrm{KL}}_{i,t}\geq
\kappa\widehat C^{\mathrm{KL}}_{i,t})
\leq e^{-2M_iD^2(1-\kappa^{-2})^2/B_i^2}
\leq\delta$, proving the claim.
\end{proof}
\color{black}
\section{Algorithm Design}
\label{sec:algorithm}

We now develop synchronization policies $\pi$ that can be evaluated
before a device installs the current server model.
Proposition~\ref{prop:aoint-control} shows that the empirical KL
certificate controls AoInt, and hence stale-model regret, with high
probability. In particular, for a fixed model pair satisfying the
probe requirement of Proposition~\ref{prop:aoint-control},
$
r_{i,t}(\theta_{i,t})
\leq C^{\mathrm{KL}}_{i,t}
<
\kappa\widehat C^{\mathrm{KL}}_{i,t}
$
with the corresponding high probability. This relation provides the
basis for the empirical synchronization certificates developed below.

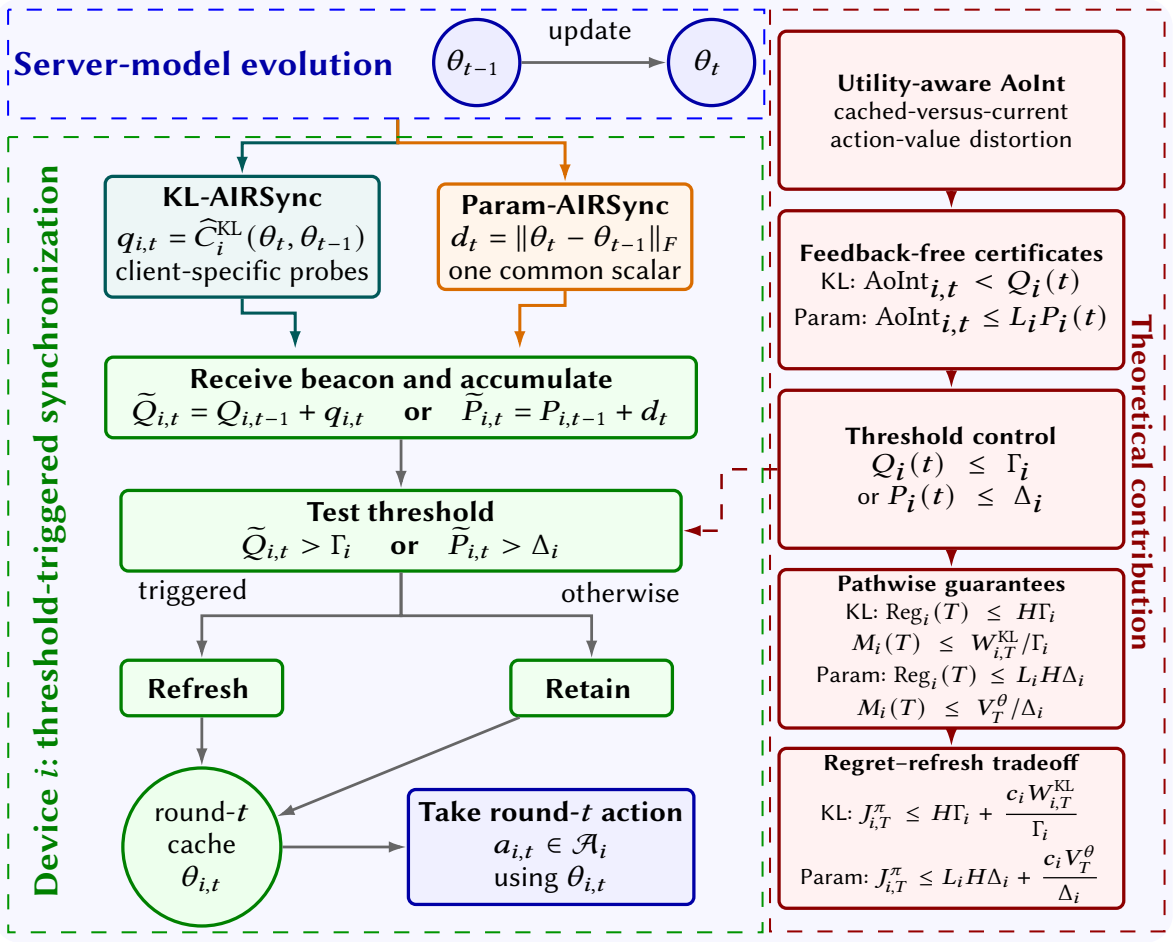
\begin{figure}
    \centering
    \scalebox{0.95}{% Required in the preamble:
% \usepackage{tikz}
% \usetikzlibrary{arrows.meta}

\begin{tikzpicture}[
    x=1mm,
    y=1mm,
    >={Latex[length=1.25mm,width=.8mm]},
    font=\sffamily\fontsize{5.8}{6.5}\selectfont
]

% --------------------------------------------------------------------------
% Background
% --------------------------------------------------------------------------
\fill[
    blue!3,
    rounded corners=2mm
] (-37.7,-67.35) rectangle (44.46,-1.57);

% --------------------------------------------------------------------------
% Server-model evolution
% --------------------------------------------------------------------------
\node[
    font=\sffamily\bfseries\footnotesize,
    text=blue!65!black
] at (-23,-6.04)
{
    Server-model evolution
};

\node[
    circle,
    draw=blue!65!black,
    fill=blue!7,
    line width=.65pt,
    minimum size=6mm,
    align=center,
    inner sep=.7pt
] (old) at (-4,-6.04)
{
    $\theta_{t-1}$
};

\node[
    circle,
    draw=blue!65!black,
    fill=blue!7,
    line width=.65pt,
    minimum size=6mm,
    align=center,
    inner sep=.7pt
] (new) at (12.35,-6.04)
{
    $\theta_t$
};

\draw[
    ->,
    line width=.55pt,
    draw=gray!80!black
] (old) -- node[above=0pt,pos=.47] {update} (new);

% --------------------------------------------------------------------------
% Server-side feedback-free beacons
% --------------------------------------------------------------------------
\node[
    rectangle,
    rounded corners=1.5pt,
    draw=teal!70!black,
    fill=teal!7,
    line width=.65pt,
    align=center,
    inner sep=2pt
] (kl) at (-20.33,-18.17)
{
    \textbf{KL-AIRSync}\\
    $q_{i,t}=\widehat{C}_i^{\mathrm{KL}}
    (\theta_t,\theta_{t-1})$\\
    client-specific probes
};

\node[
    rectangle,
    rounded corners=1.5pt,
    draw=orange!85!black,
    fill=orange!8,
    line width=.65pt,
    align=center,
    inner sep=2pt
] (param) at (2.15,-18.17)
{
    \textbf{Param-AIRSync}\\
    $d_t=\lVert\theta_t-\theta_{t-1}\rVert_F$\\
    one common scalar
};

\draw[
    ->,
    line width=.6pt,
    draw=teal!70!black
] (-9.5,-10) -- ++(0,-1.63) -| (kl.north);

\draw[
    ->,
    line width=.6pt,
    draw=orange!85!black
] (-9.5,-10) -- ++(0,-1.63) -| (param.north);

% --------------------------------------------------------------------------
% Device-side synchronization procedure
% --------------------------------------------------------------------------
\node[
    font=\sffamily\bfseries\footnotesize,
    text=green!50!black,
    rotate=90
] at (-33.74,-38.65)
{
    Device $i$: threshold-triggered synchronization
};

\node[
    rectangle,
    rounded corners=1.5pt,
    draw=green!50!black,
    fill=green!6,
    line width=.65pt,
    minimum width=117pt,
    align=center,
    inner sep=2pt
] (acc) at (-9.27,-29.36)
{
    \textbf{Receive beacon and accumulate}\\
    $\widetilde Q_{i,t}=Q_{i,t-1}+q_{i,t}$
    \quad\textbf{or}\quad
    $\widetilde P_{i,t}=P_{i,t-1}+d_t$
};

\draw[
    ->,
    line width=.6pt,
    draw=teal!70!black
] (kl.south) -- ++(0,-1) -| (-16.64,-26);

\draw[
    ->,
    line width=.6pt,
    draw=orange!85!black
] (param.south) -- ++(0,-1) -| (-1,-26);

\node[
    rectangle,
    rounded corners=1.5pt,
    draw=green!50!black,
    fill=green!6,
    line width=.65pt,
    minimum width=39mm,
    align=center,
    inner sep=2pt
] (test) at (-9.27,-38.64)
{
    \textbf{Test threshold}\\
    $\widetilde Q_{i,t}>\Gamma_i$
    \quad\textbf{or}\quad
    $\widetilde P_{i,t}>\Delta_i$
};

\draw[
    ->,
    line width=.55pt,
    draw=gray!80!black
] (acc) -- (test);

\node[
    rectangle,
    rounded corners=1.5pt,
    draw=green!50!black,
    fill=green!6,
    line width=.65pt,
    minimum width=32pt,
    minimum height=11pt,
    align=center,
    inner sep=2pt
] (refresh) at (-23.14,-49.59)
{
    \textbf{Refresh}
};

\node[
    rectangle,
    rounded corners=1.5pt,
    draw=green!50!black,
    fill=green!6,
    line width=.65pt,
    minimum width=32pt,
    minimum height=11pt,
    align=center,
    inner sep=2pt
] (retain) at (4,-49.59)
{
    \textbf{Retain}
};

\draw[
    ->,
    line width=.55pt,
    draw=gray!80!black
] (test.south) -- ++(0,-3) -| (refresh.north);

\draw[
    ->,
    line width=.55pt,
    draw=gray!80!black
] (test.south) -- ++(0,-3) -| (retain.north);

\node at (-23.76,-42.93)
{
    triggered
};

\node at (6.01,-42.93)
{
    otherwise
};

\node[
    circle,
    draw=green!50!black,
    fill=green!7,
    line width=.65pt,
    minimum size=11mm,
    align=center,
    inner sep=.7pt
] (cache) at (-23.14,-60.66)
{
    round-$t$\\
    cache\\
    $\theta_{i,t}$
};

\node[
    rectangle,
    rounded corners=1.5pt,
    draw=blue!65!black,
    fill=blue!6,
    line width=.65pt,
    minimum height=8mm,
    align=center,
    inner sep=2pt
] (action) at (1.18,-60.66)
{
    \textbf{Take round-$t$ action}\\
    $a_{i,t}\in\mathcal A_i$\\
    using $\theta_{i,t}$
};

\draw[
    ->,
    line width=.55pt,
    draw=gray!80!black
] (refresh) -- (cache);

\draw[
    ->,
    line width=.55pt,
    draw=gray!80!black
] (retain) -- (cache);

\draw[
    ->,
    line width=.55pt,
    draw=gray!80!black
] (cache) -- (action);

% --------------------------------------------------------------------------
% Theoretical contribution
% --------------------------------------------------------------------------
\node[font=\sffamily\bfseries\scriptsize, text=red!55!black, rotate=-90] at (42.35,-35.3)
{
    Theoretical contribution
};

% All theoretical boxes have the same height, and their centers are
% separated by 12.5 mm to ensure equal vertical spacing.
\node[
    rectangle,
    rounded corners=1.5pt,
    draw=red!55!black,
    fill=red!5,
    line width=.65pt,
    minimum width=24mm,
    minimum height=11mm,
    text width=22mm,
    align=center,
    inner sep=1pt,
    font=\sffamily\fontsize{5}{5.6}\selectfont
] (aoint) at (29.05,-9.36)
{
    \textbf{Utility-aware AoInt}\\
    cached-versus-current\\
    action-value distortion
};

\node[
    rectangle,
    rounded corners=1.5pt,
    draw=red!55!black,
    fill=red!5,
    line width=.65pt,
    minimum width=24mm,
    minimum height=11mm,
    text width=22mm,
    align=center,
    inner sep=1pt,
    font=\sffamily\fontsize{5}{5.6}\selectfont
] (cert) at (29.05,-21.86)
{
    \textbf{Feedback-free certificates}\\
    KL: $\operatorname{AoInt}_{i,t}< Q_i(t)$\\
    Param: $\operatorname{AoInt}_{i,t}
    \leq L_iP_i(t)$
};

\draw[
    ->,
    line width=.6pt,
    draw=red!55!black
] (aoint) -- (cert);

\node[
    rectangle,
    rounded corners=1.5pt,
    draw=red!55!black,
    fill=red!5,
    line width=.65pt,
    minimum width=24mm,
    minimum height=11mm,
    text width=22mm,
    align=center,
    inner sep=1pt,
    font=\sffamily\fontsize{5}{5.6}\selectfont
] (threshold-control) at (29.05,-34.36)
{
    \textbf{Threshold control}\\
    $Q_i(t)\leq\Gamma_i$\\
    or $P_i(t)\leq\Delta_i$
};

\draw[
    ->,
    line width=.6pt,
    draw=red!55!black
] (cert) -- (threshold-control);

\node[
    rectangle,
    rounded corners=1.5pt,
    draw=red!55!black,
    fill=red!5,
    line width=.65pt,
    minimum width=24mm,
    minimum height=11mm,
    text width=22mm,
    align=center,
    inner sep=1pt,
    font=\sffamily\fontsize{4.7}{5.2}\selectfont
] (bounds) at (29.05,-46.86)
{
    \textbf{Pathwise guarantees}\\
    KL: $\operatorname{Reg}_i(T)\leq H\Gamma_i$\\
    $M_i(T)\leq W_{i,T}^{\mathrm{KL}}/\Gamma_i$\\
    Param: $\operatorname{Reg}_i(T)\leq L_iH\Delta_i$\\
    $M_i(T)\leq V_T^\theta/\Delta_i$
};

\draw[
    ->,
    line width=.6pt,
    draw=red!55!black
] (threshold-control) -- (bounds);

\node[
    rectangle,
    rounded corners=1.5pt,
    draw=red!55!black,
    fill=red!5,
    line width=.65pt,
    minimum width=24mm,
    minimum height=11mm,
    text width=22mm,
    align=center,
    inner sep=1pt,
    font=\sffamily\fontsize{4.6}{5.1}\selectfont
] (joint-cost) at (29.05,-59.36)
{
    \textbf{Regret--refresh tradeoff}\\
    KL: $J_{i,T}^{\pi}\leq
    H\Gamma_i+
    \dfrac{c_iW_{i,T}^{\mathrm{KL}}}{\Gamma_i}$\\
    Param: $J_{i,T}^{\pi}\leq
    L_iH\Delta_i+
    \dfrac{c_iV_T^\theta}{\Delta_i}$
};

\draw[
    ->,
    line width=.6pt,
    draw=red!55!black
] (bounds) -- (joint-cost);

% --------------------------------------------------------------------------
% Link the algorithmic threshold to its theoretical implication
% --------------------------------------------------------------------------
\draw[
    ->,
    dashed,
    line width=.5pt,
    draw=red!55!black
] (threshold-control.west) -- ++(-3.89,0) |- (test.east);

  \draw[dashed, blue] (-36.63,-2.36) rectangle (16.02,-9.93);
  \draw[dashed, green!58!black] (-36.62,-11.23) rectangle (15.9,-66.6);

  \draw[dashed, red!59!black] (16.42,-2.36) rectangle (43.89,-66.54);
\end{tikzpicture}}
    \caption{An illustration of Algorithm 1 in an edge-AI system.}
    \label{fig:ill}
\end{figure}

Direct evaluation of the endpoint discrepancy, however, requires
comparing the current server model with the device's cached version.
We therefore begin with \klairsync, which accumulates output-space
changes between consecutive server models, and then introduce
\airsync, which replaces client-specific output comparisons with
$\lVert\theta_t-\theta_{t-1}\rVert_F$.
Both mechanisms use the same timing. At $t=1$, device $i$ is
provisioned with $\theta_1$, initializes its accumulator to zero, and
acts with $\theta_{i,1}=\theta_1$. For each $t\geq2$, it inherits the
previous cache, receives the relevant beacon, updates the accumulator
tentatively, tests the threshold, and, if triggered, installs
$\theta_t$ and resets the accumulator $Q_i,P_i=0$ before taking its
round-$t$ action. Otherwise, it retains the inherited cache
$\theta_{i,t}$ and the updated accumulator. An overview is provided
in Fig.~\ref{fig:ill}.

\subsection{Kullback--Leibler-Aware Age-of-Intelligence Synchronization}

The key design parameter governing Algorithm~\ref{alg:airsync} is its
synchronization threshold. Following
Proposition~\ref{prop:aoint-control}, we define the KL-based change
between consecutive server models as
$
q_{i,t}:=
\kappa\widehat C_{i,t}^{\mathrm{KL}}.
\label{eq:kl-beacon}
$
For each adjacent model pair, Proposition~\ref{prop:aoint-control}
implies that
$
C_i^{\mathrm{KL}}(\theta_t,\theta_{t-1})
<
q_{i,t}
\label{eq:adjacent-high-prob}
$
with high probability whenever its probe requirement is satisfied.

To obtain a horizon-wide guarantee, let $H:=T-1$ and apply
Proposition~\ref{prop:aoint-control} to every adjacent model pair with
failure probability $\delta/H$. Accordingly, it is sufficient that
the probe requirement in Proposition~\ref{prop:aoint-control} be
satisfied with $\delta$ replaced by $\delta/H$. In particular, for
each adjacent model pair, the corresponding requirement becomes
$
M_i
\geq
B_i^2\log(H/\delta)/
\left[2D^2(1-\kappa^{-2})^2\right].
\label{eq:horizon-probe-requirement}
$
A union bound then ensures that all adjacent empirical certificates
simultaneously dominate their population counterparts with probability
at least $1-\delta$.

In practice, each client $i$ provides a fixed unlabeled probe set
$\mathcal S$ once. The server evaluates each pair of adjacent models
on this set and unicasts the resulting scalar $q_{i,t}$. The signal
is client-specific through the client's context distribution, yet it
requires neither labels nor observed utilities. To prevent a sequence
of individually small changes from going undetected, device $i$
accumulates the adjacent certificates since its most recent
synchronization:
$
Q_i(t):=
\sum\nolimits_{\ell=\tau_{i,t}+1}^{t}q_{i,\ell}.
\label{eq:cumulative-Q}
$
Because each $q_{i,\ell}$ is an empirical high-probability upper
certificate rather than a deterministic population certificate, the
resulting control of AoInt is itself a high-probability statement.
The following lemma makes this relation explicit.

\begin{lemma}[Cumulative KL Certificate]
\label{lem:cumulative}
Under a fixed context distribution $\mu_i$ and action set
$\mathcal A_i$, suppose that the probe requirement of
Proposition~\ref{prop:aoint-control} holds for each adjacent model
pair after the most recent synchronization. \color{black}Define
$
\mathcal E_{i,t}:=
\bigcap_{\ell=\tau_{i,t}+1}^{t}
\left\{
C_i^{\mathrm{KL}}(\theta_\ell,\theta_{\ell-1})
<
q_{i,\ell}
\right\}.
\label{eq:certificate-event}
$
\color{black}Then, on $\mathcal E_{i,t}$,
$
\AoInt_{i,t}
=
r_{i,t}(\theta_{i,t})
<
Q_i(t).
\label{eq:cumulative-kl-control}
$
Moreover, if each adjacent certificate has failure probability at
most $\delta/H$, then
\begin{equation}
\Pr(\mathcal E_{i,t})
\geq
1-(t-\tau_{i,t})\delta/H
\geq
1-\delta.
\label{eq:cumulative-event-prob}
\end{equation}
\end{lemma}

\begin{proof}
Fix $t$ and write $\tau:=\tau_{i,t}$. The difference between the
current and cached expected utilities can be decomposed along the
server-model path from $\theta_\tau$ to $\theta_t$. Applying the
triangle inequality gives
$
\AoInt_{i,t}
=
r_{i,t}(\theta_{i,t})
\leq
\sum\nolimits_{\ell=\tau+1}^{t}
C_i^{\mathrm{KL}}(\theta_\ell,\theta_{\ell-1}).
\label{eq:cumulative-population}
$
On the event $\mathcal E_{i,t}$,
Proposition~\ref{prop:aoint-control} gives
$
C_i^{\mathrm{KL}}(\theta_\ell,\theta_{\ell-1})
<
q_{i,\ell}
=
\kappa\widehat C_{i,\ell}^{\mathrm{KL}}
$
for every $\ell=\tau+1,\ldots,t$. Therefore,
$
\AoInt_{i,t}
=
r_{i,t}(\theta_{i,t})
\leq
\sum\nolimits_{\ell=\tau+1}^{t}
C_i^{\mathrm{KL}}(\theta_\ell,\theta_{\ell-1})
<
Q_i(t).
$
Finally, each adjacent-pair certificate fails with probability at
most $\delta/H$. Applying the union bound over
$t-\tau_{i,t}\leq H$ adjacent model pairs gives
$
\Pr(\mathcal E_{i,t})
\geq
1-(t-\tau_{i,t})\delta/H
\geq
1-\delta.
$
\end{proof}

\begin{algorithm}[t]
\caption{Certificate-Based AIRSync at device $i$}
\label{alg:airsync}
\begin{algorithmic}[1]
\Require Initial model $\theta_1$ and threshold $\Gamma_i$ for
\klairsync or $\Delta_i$ for \airsync
\State $z_{i,1}\gets0$, $\theta_{i,1}\gets\theta_1$, and
$\tau_{i,1}\gets1$
\State $Q_i\gets0$ for \klairsync \textbf{or}
$P_i\gets0$ for \airsync
\State Observe $s_{i,1}$ and select $a_{i,1}$ using $\theta_{i,1}$
\For{$t=2,\ldots,T$}
    \State $z_{i,t}\gets0$, $\theta_{i,t}\gets\theta_{i,t-1}$,
    and $\tau_{i,t}\gets\tau_{i,t-1}$
    \State Receive
    $q_{i,t}=\kappa\widehat C_i^{\mathrm{KL}}
    (\theta_t,\theta_{t-1})$
    \textbf{or}
    $d_t=\lVert\theta_t-\theta_{t-1}\rVert_F$
    \State $Q_i\gets Q_i+q_{i,t}$
    \textbf{or}
    $P_i\gets P_i+d_t$
    \If{$Q_i>\Gamma_i$ \textbf{or} $P_i>\Delta_i$}
        \State $z_{i,t}\gets1$, $\theta_{i,t}\gets\theta_t$,
        and $\tau_{i,t}\gets t$
        \State $Q_i\gets0$ \textbf{or} $P_i\gets0$
    \EndIf
    \State Observe $s_{i,t}$ and select $a_{i,t}$ using
    $\theta_{i,t}$
\EndFor
\end{algorithmic}
\end{algorithm}

We next establish the regret--refresh guarantee for the \klairsync
instance of Algorithm~\ref{alg:airsync}. Let $H:=T-1$ and
$
W_{i,T}^{\mathrm{KL}}
:=
\sum\nolimits_{t=2}^{T}q_{i,t}
$
denote the total empirical predictive variation after initialization.
Let $\Gamma_i\in\mathbb R_{+}$ be the device-specific
cumulative-certificate threshold. \color{black}Define the horizon-wide event
$
\mathcal E_i:=
\bigcap_{\ell=2}^{T}
\left\{
C_{i,t}^{\mathrm{KL}}
<
q_{i,\ell}
\right\}.
\label{eq:horizon-event}
$ 
\color{black}Under \eqref{eq:horizon-probe-requirement}, the union bound gives
$
\Pr(\mathcal E_i)\geq1-\delta.
\label{eq:horizon-event-prob}
$

On $\mathcal E_i$, the recursion above ensures that $Q_i(t)$ controls
the stale-model regret accumulated since the latest refresh.
Moreover, $Q_i(t)\leq\Gamma_i$ on every non-refresh round. Thus,
smaller thresholds provide tighter high-probability control of
stale-model regret but generally induce more refreshes. For each
accepted $q_{i,\ell}$, let $\rho_{i,\ell}$ count the subsequent
non-refresh rounds, including round $\ell$, in which it remains in
$Q_i$; set $\rho_{i,\ell}=0$ if it triggers a refresh.

\begin{theorem}[Residence-Time Regret--Refresh Tradeoff]
\label{thm:tradeoff}
Let $T\geq2$, $\Gamma_i>0$, and fix $\mu_i$, $\mathcal A_i$, and
$h_i$ over the horizon. Suppose that the probe requirement
\eqref{eq:horizon-probe-requirement} holds for every adjacent model
pair. Then, \color{black}with probability at least $1-\delta$, \color{black} \klairsync satisfies
\begin{subequations}
\label{eq:pathwise-bounds}
\begin{align}
&\Reg_i(T)
<
\sum\nolimits_{\ell=2}^{T}
q_{i,\ell}\rho_{i,\ell}
\leq
H\Gamma_i,
\label{subeq:regret}\\
&M_i(T)
\leq
\max\nolimits\!\left\{
0,
\left\lceil
W_{i,T}^{\mathrm{KL}}/\Gamma_i
\right\rceil-1
\right\}
\leq
W_{i,T}^{\mathrm{KL}}/\Gamma_i,
\\
&\Reg_i(T)+c_iM_i(T)
<
H\Gamma_i+
c_iW_{i,T}^{\mathrm{KL}}/\Gamma_i.
\end{align}
\end{subequations}
The residence-time bound may be sharper than $H\Gamma_i$ when changes
have short residence times or the accumulated certificate is usually
well below the threshold.
\end{theorem}

The first bound shows that, on the high-probability event
$\mathcal E_i$, each accepted predictive change contributes in
proportion to its residence time in the accumulator. A change followed
quickly by synchronization therefore affects few decisions, whereas a
persistent change contributes repeatedly. The coarser bound
$H\Gamma_i$ follows because the accumulator remains below $\Gamma_i$
on every non-refresh round. The refresh-count bound, in contrast,
follows directly from the deterministic threshold recursion applied
to the observed certificates $q_{i,t}$. Each synchronization requires
more than $\Gamma_i$ accumulated empirical predictive variation.
Thus, decreasing the threshold provides tighter high-probability
control of stale-decision regret but increases refreshes,
while increasing it has the opposite effect.

\begin{proof}
Consider the event $\mathcal E_i$. By
Lemma~\ref{lem:cumulative}, on every non-refresh round,
$
\AoInt_{i,t}
<
Q_i(t).
$
On a refresh round, $\theta_t$ is cached before acting and hence
$\AoInt_{i,t}=r_{i,t}(\theta_t)=0$. Therefore, on
$\mathcal E_i$,
$
\Reg_i(T)
<
\sum\nolimits_{t:z_{i,t}=0}Q_i(t)
=
\sum\nolimits_{\ell=2}^{T}
q_{i,\ell}\rho_{i,\ell}.
\label{eq:residence-proof}
$
The equality counts each accepted certificate once for every
non-refresh round in which it remains in the accumulator. Because
$Q_i(t)\leq\Gamma_i$ on every such round and there are at most
$H$ post-initialization rounds,
$
\Reg_i(T)<H\Gamma_i.
$

For the refresh count, let $t_1<\cdots<t_m$ be the
$m=M_i(T)$ refresh rounds and set $t_0=1$. A refresh at $t_j$
requires
$
\sum_{\ell=t_{j-1}+1}^{t_j}q_{i,\ell}>\Gamma_i.
$
These triggering segments are disjoint. Summing them gives
$
m\Gamma_i<W_{i,T}^{\mathrm{KL}}.
$
Since $m$ is a nonnegative integer,
$
m
\leq
\max\left\{
0,
\left\lceil
W_{i,T}^{\mathrm{KL}}/\Gamma_i
\right\rceil-1
\right\}.
$
This refresh-count statement follows directly from the observed
certificate sequence and does not require $\mathcal E_i$.
Combining it with the regret bound on $\mathcal E_i$ gives
$
\Reg_i(T)+c_iM_i(T)
<
H\Gamma_i+
c_iW_{i,T}^{\mathrm{KL}}/\Gamma_i.
$
Finally, \eqref{eq:horizon-event-prob} gives
$\Pr(\mathcal E_i)\geq1-\delta$, completing the proof.
\end{proof}

\begin{corollary}[Envelope-Optimal Fixed Threshold]
\label{cor:threshold}
Suppose
$0<W_{i,T}^{\mathrm{KL}}\leq\bar W_{i,T}$ is known before operation.
Under the conditions of Theorem~\ref{thm:tradeoff}, among fixed
thresholds, the choice
$
\Gamma_i^*
=
\sqrt{c_i\bar W_{i,T}/H}
\label{eq:optimal-kl-threshold}
$
uniquely minimizes the corresponding worst-case envelope and, with
\color{black}probability \color{black} at least $1-\delta$, guarantees
\begin{subequations}
\begin{align}
&\Reg_i(T)
<
\sqrt{c_iH\bar W_{i,T}},\quad
M_i(T)
<
\sqrt{H\bar W_{i,T}/c_i},
\\
&\Reg_i(T)+c_iM_i(T)
<
2\sqrt{c_iH\bar W_{i,T}}.
\label{eq:kl-optimal-envelope}
\end{align}
\end{subequations}

For $c_i=\Theta(1)$,
$\bar W_{i,T}=\mathcal O(T)$ gives an
$\mathcal O(T)$ cumulative-regret bound, while
$\bar W_{i,T}=o(T)$ \color{black}ensures sublinear cumulative regret, and hence vanishing average regret.\color{black}

%that the regret vanishes.
\end{corollary}

This corollary makes the threshold tradeoff explicit. The optimal
fixed threshold follows from balancing the high-probability
stale-model regret envelope against the synchronization cost.

\begin{proof}
Let
$
g(\Gamma)
:=
H\Gamma+
c_i\bar W_{i,T}/\Gamma
\quad(\Gamma>0).
$
Since
$
g''(\Gamma)
=
2c_i\bar W_{i,T}/\Gamma^3>0,
$
$g$ is strictly convex. Solving $g'(\Gamma)=0$ gives the
minimizer
$
\Gamma_i^*
=
\sqrt{c_i\bar W_{i,T}/H}.
$
Substituting this value into the bounds of
Theorem~\ref{thm:tradeoff} gives the stated results. Moreover,
$
\Reg_i(T)/T
<
\sqrt{
c_i(H/T)(\bar W_{i,T}/T)
},
$
converging to $0$ when
$\bar W_{i,T}=o(T)$.
\end{proof}

\subsection{Parameter-Aware Age-of-Intelligence Synchronization}

Although \klairsync directly tracks client-specific predictive
changes, evaluating adjacent models on a separate probe set for every
device introduces additional inference, communication, and storage
overhead that may become substantial in large-scale,
resource-constrained IoT systems. To obtain a lighter synchronization
mechanism, we therefore relate the KL certificate to parameter changes
measured by the fixed Frobenius norm $\lVert\cdot\rVert_F$. This
relation enables \airsync to replace client-specific probe inference
with a single scalar parameter-change signal computed by the server
and shared across devices. We first specify the conditions under which
the proposed algorithm is applied.

\begin{assumption}
\label{ass:smooth}
Along the model path of interest, there is a finite $L_i$ such that
$
C_i^{\mathrm{KL}}(\theta,\theta')
\leq
L_i\lVert\theta-\theta'\rVert_F.
$
\end{assumption}

Assumption~\ref{ass:smooth} does not imply that every parameter change
affects the device's decisions. Instead, it provides a conservative
sensitivity bound along the relevant model path. Unlike the empirical
KL certificate used by \klairsync, this condition directly controls
the population certificate and therefore does not introduce
finite-probe estimation uncertainty.

The following lemma verifies this condition for a softmax model, where
$W$ and $W'$ are two weight matrices in
$\mathbb{R}^{|\mathcal{Y}|\times|\mathcal{Y}|}$ and
$\lVert\cdot\rVert_F$ denotes the Frobenius norm, with
$
\lVert W\rVert_F
:=
\sqrt{
\sum_{j=1}^{|\mathcal{Y}|}
\sum_{k=1}^{|\mathcal{Y}|}
W_{jk}^{2}
}.
$

\begin{lemma}[Softmax Parameter-to-KL Smoothness]
\label{lem:softmax-smoothness}
Let
$p_W(\cdot\mid s)=\operatorname{softmax}(W^\top\phi(s))$, where
$\phi:\mathcal S\to\mathbb R^d$ is the feature representation used
by the softmax layer. Furthermore, we apply a fixed $\ell_2$
normalization before the softmax layer:
$
\phi(s)
=
R_i\widetilde{\phi}(s)/
\max\{\lVert\widetilde{\phi}(s)\rVert_2,R_i\},
$
where $\widetilde{\phi}(s)$ denotes the original feature vector.
We set $R_i$ to the prescribed feature-norm bound used in this
normalization and consequently use
$L_i=R_i/(2\sqrt{2})$ in \airsync. Suppose that
$\lVert\phi(s)\rVert_2\leq R_i$ for every $s$ in the support of
$\mu_i$. Then
$
C_i^{\mathrm{KL}}(W,W')
\leq
R_i\lVert W-W'\rVert_F/(2\sqrt2).
\label{eq:softmax-param-kl}
$
Thus, the smoothness condition holds under the Frobenius norm
with $L_i=R_i/(2\sqrt2)$.
\end{lemma}

\begin{proof}
Let
$A(z):=\log\sum_y e^{z_y}$ and
$[p_z]_y:=e^{z_y}/\sum_{y'}e^{z_{y'}}$. Then
$\nabla A(z)=p_z$ and
$
\nabla^2A(z)
=
\operatorname{diag}(p_z)-p_zp_z^\top.
$
For any $u$,
$
u^\top\nabla^2A(z)u
=
\operatorname{Var}_{Y\sim p_z}(u_Y).
$
Popoviciu's variance bound and
$\max_yu_y-\min_yu_y\leq\sqrt2\lVert u\rVert_2$ imply
$
u^\top\nabla^2A(z)u
\leq
(1/2)\lVert u\rVert_2^2;
$
hence, $A$ is $1/2$-smooth under the Euclidean norm.
For a fixed $s$, let
$z=W^\top\phi(s)$ and $z'=W'^\top\phi(s)$. Expressing the categorical
KL divergence as the corresponding Bregman divergence of $A$ gives
$
\KL\!\left(
p_W(\cdot\mid s)
\middle\|
p_{W'}(\cdot\mid s)
\right)
\leq
(1/4)\lVert z-z'\rVert_2^2
\leq
R_i^2\lVert W-W'\rVert_F^2/4.
$
Taking expectation over $s\sim\mu_i$, multiplying by $1/2$, and
taking the square root yields
$
C_i^{\mathrm{KL}}(W,W')
\leq
R_i\lVert W-W'\rVert_F/(2\sqrt2).
$
In the experiments, we use the
fixed normalization
$
\phi(s)
=
R_i\widetilde{\phi}(s)/
\max\{\lVert\widetilde{\phi}(s)\rVert_2,R_i\},
$
where $\widetilde{\phi}(s)$ is the original feature vector. Therefore,
$\lVert\phi(s)\rVert_2\leq R_i$, with given $R_i$ and
$L_i=R_i/(2\sqrt2)$.
\end{proof}

\begin{proposition}[Parameter Variation Controls AoInt and Regret]
\label{prop:parameter}
Let
$
d_t:=
\lVert\theta_t-\theta_{t-1}\rVert_F
$
and
$
P_i(t):=
\sum\nolimits_{\ell=\tau+1}^{t}d_\ell.
$
Under Assumption~\ref{ass:smooth},
$
\AoInt_{i,t}
=
r_{i,t}(\theta_{i,t})
\leq
L_iP_i(t).
\label{eq:param-regret}
$
\end{proposition}

$P_i(t)$ is the parameter-space path length accumulated while device
$i$ retains its cached model. Because successive parameter changes may
cancel or occur in directions that barely affect model outputs, this
certificate can be conservative. Nevertheless, under
Assumption~\ref{ass:smooth}, it provides a deterministic upper bound
on AoInt and stale-model regret and does not require finite-probe
estimation.

\begin{proof}
Fix $t$ and let $\tau:=\tau_{i,t}$. By the same path decomposition
used in Lemma~\ref{lem:cumulative},
$
\AoInt_{i,t}
\leq
\sum_{\ell=\tau+1}^{t}
C_i^{\mathrm{KL}}(\theta_\ell,\theta_{\ell-1}).
$
Applying Assumption~\ref{ass:smooth} to each adjacent pair gives
$
\AoInt_{i,t}
\leq
L_i
\sum_{\ell=\tau+1}^{t}
\lVert\theta_\ell-\theta_{\ell-1}\rVert_F
=
L_iP_i(t),
$
which proves the result.
\end{proof}

We can now obtain the analogous regret--refresh tradeoff for the
\airsync instance of Algorithm~\ref{alg:airsync} by replacing
predictive variation with accumulated parameter variation. Unlike the
KL result, the following bound is deterministic under
Assumption~\ref{ass:smooth} and therefore does not require the
high-probability event $\mathcal E_i$.

\begin{corollary}[\airsync Regret--Refresh Tradeoff]
\label{cor:param-trigger}
Let
$
V_T^\theta
:=
\sum\nolimits_{t=2}^{T}d_t,
$
and let $\sigma_{i,\ell}$ count the non-refresh rounds for which an
accepted $d_\ell$ remains in $P_i$. Define
$
S_i:=
\sum\nolimits_{\ell=2}^{T}
d_\ell\sigma_{i,\ell}.
$
Then the \airsync instance of Algorithm~\ref{alg:airsync} satisfies
\begin{subequations}
\begin{align}
&\Reg_i(T)
\leq
L_iS_i
\leq
L_iH\Delta_i,
\label{eq:param-regret-tradeoff}\quad
M_i(T)
\leq
V_T^\theta/\Delta_i,
\\
&\Reg_i(T)+c_iM_i(T)
\leq
L_iH\Delta_i+
c_iV_T^\theta/\Delta_i.
\label{eq:param-tradeoff}
\end{align}
\end{subequations}
If
$V_T^\theta\leq\bar V_T^\theta$, the last envelope is minimized by
$
\Delta_i^*
=
\sqrt{c_i\bar V_T^\theta/(L_iH)},
\label{eq:optimal-param-threshold}
$
and is at most
$
2\sqrt{
L_ic_iH\bar V_T^\theta
}.
$
Hence, constant $c_i,L_i$ and
$\bar V_T^\theta=o(T)$ imply sublinear cumulative stale-model regret.
\end{corollary}

This result mirrors the \klairsync tradeoff, with the parameter-path
budget $V_T^\theta$ replacing predictive variation. The sensitivity
constant $L_i$ maps the server model's common parameter motion to a
client-specific deterministic bound on output variation.

\begin{proof}
Let $\mathcal N_i$ be the non-refresh rounds after initialization.
For $t\in\mathcal N_i$,
Proposition~\ref{prop:parameter} gives
$
\AoInt_{i,t}
=
r_{i,t}(\theta_{i,t})
\leq
L_iP_i(t),
$
whereas refresh rounds have zero stale-model regret. Summing over
$\mathcal N_i$ and exchanging the two finite sums yields
$
\Reg_i(T)
\leq
L_i\sum\nolimits_{t\in\mathcal N_i}P_i(t)
=
L_i
\sum\nolimits_{\ell=2}^{T}
d_\ell\sigma_{i,\ell}.
$
The equality counts each accepted increment $d_\ell$ once for every
non-refresh round in which it remains in $P_i$. Since
$P_i(t)\leq\Delta_i$ and $|\mathcal N_i|\leq H$,
$
\Reg_i(T)
\leq
L_iH\Delta_i.
$

For the refresh count, write the refresh rounds as
$t_1<\cdots<t_m$ and set $t_0=1$. Each triggering segment contains
more than $\Delta_i$ path length, and the segments are disjoint.
Hence,
$
m\Delta_i<V_T^\theta,
$
which gives
$
M_i(T)
\leq
V_T^\theta/\Delta_i.
$
Adding $c_iM_i(T)$ gives
$
\Reg_i(T)+c_iM_i(T)
\leq
L_iH\Delta_i+
c_iV_T^\theta/\Delta_i.
$
With $V_T^\theta\leq\bar V_T^\theta$, define
$
f(\Delta)
=
L_iH\Delta+
c_i\bar V_T^\theta/\Delta.
$
Since
$
f''(\Delta)
=
2c_i\bar V_T^\theta/\Delta^3>0,
$
$f$ is strictly convex. Solving $f'(\Delta)=0$ gives
$
\Delta_i^*
=
\sqrt{c_i\bar V_T^\theta/(L_iH)}.
$
Substitution gives
$
\Reg_i(T)+c_iM_i(T)
\leq
2\sqrt{
L_ic_iH\bar V_T^\theta
}.
$
Finally,
$
\Reg_i(T)/T
\leq
\sqrt{
L_ic_i(H/T)(\bar V_T^\theta/T)
}
\longrightarrow0,
\qquad
\bar V_T^\theta=o(T).
$
\end{proof}

\section{Numerical Results}
\color{black}
\begin{figure*}[t]
\centering
\includegraphics[width=\textwidth]{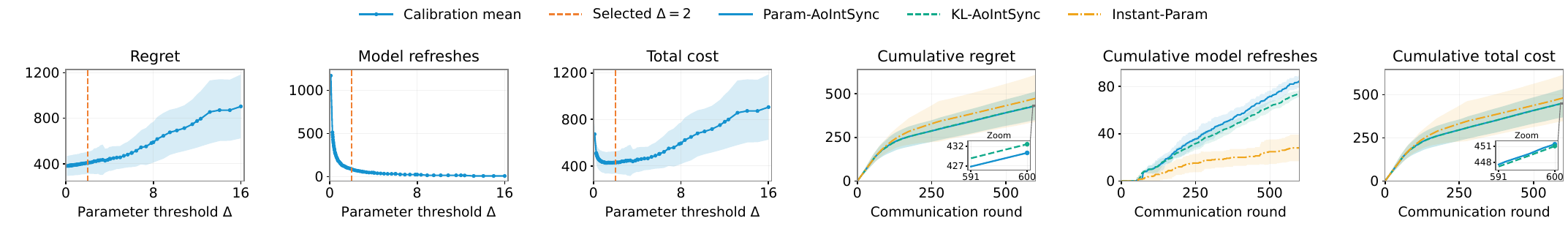}
\vspace{-0.6cm}
\caption{Regret and refresh-cost tradeoff.
    The first three panels show \airsync's calibration results
    as its threshold varies; the vertical dashed line marks the
    selected threshold $\Delta=2$.
    The last three panels compare cumulative regret, refresh count,
    and total cost over 20 fresh validation traces.
    Main-panel shading denotes one sample standard deviation.
    }
    \vspace{-0.2cm}
% \airsync 
%\Description{The top row shows regret, model refreshes, and total cost as the parameter threshold varies. The bottom row compares cumulative regret, model refreshes, and total cost across policies.}
%\label{fig:threshold-suite}
%\label{fig:cumulative-suite}
\label{fig:action-value-results}
\end{figure*}

\begin{table}[t]
\caption{Performance comparison on 20 fresh validation traces.
Values are mean $\pm$ standard deviation.}
\label{tab:numerical-results}
\centering
\small
\setlength{\tabcolsep}{4pt}
\begin{tabular}{lccc}
\toprule
Method & Regret & Refreshes & Total Cost \\
\midrule

\airsync
& $\mathbf{430.32 \pm 82.66}$
& $\mathit{84.40 \pm 5.49}$
& $\mathit{451.42 \pm 83.08}$ \\

\klairsync
& $432.53 \pm 80.71$
& $74.20 \pm 2.67$
& $\mathbf{451.08 \pm 80.99}$ \\

Instant-Parameters
& $\mathit{474.02 \pm 135.03}$
& $\mathbf{28.00 \pm 11.16}$
& $481.02 \pm 134.19$ \\
\bottomrule
\end{tabular}
\end{table}

%We evaluate whether the proposed methods realize the predicted regret--refresh tradeoff and whether cumulative model changes improve synchronization. We compare Param-AoIntSync and KL-AoIntSync with \emph{Instant-Parameters} policy, which refreshes when $\|\theta_t-\theta_{t-1}\|_F>\delta$ without accumulating sub-threshold changes.

We evaluate the regret--refresh tradeoff and the benefit of accumulating model changes by comparing \airsync and \klairsync with Instant-Parameters policy, which refreshes when $\|\theta_t-\theta_{t-1}\|_F>\delta$ without accumulating sub-threshold changes.

We consider an exogenous three-class softmax model over $T=600$
communication rounds and $N=8$ heterogeneous clients. The server
model evolves through gradual Gaussian parameter drift with standard
deviation $0.006$ together with six abrupt changes of Frobenius norm
$1.6$. Each client has a finite population of $1024$ contexts and
$96$ fixed unlabeled probes. The refresh cost is $c_i=0.25$.
The synchronization thresholds are selected using calibration traces
and then frozen before evaluation on $20$ fresh validation traces.
%Thus, none of the reported validation results is used for threshold selection.

We first compare the overall performance of the proposed cumulative
synchronization policies with the non-accumulating Instant-Param
baseline. Table~\ref{tab:numerical-results} reports the mean and
standard deviation of regret, number of model refreshes, and total
cost over 20 fresh validation traces. As shown in Table~\ref{tab:numerical-results}, \airsync and
\klairsync achieve mean regrets of $430.32$ and $432.53$,
respectively, compared with $474.02$ for Instant-Parameters. Their
corresponding total costs are $451.42$ and $451.08$, both lower than
the $481.02$ obtained by Instant-Parameters. This improvement is achieved
at the expense of more frequent synchronization: \airsync and
\klairsync perform $84.40$ and $74.20$ refreshes on average,
respectively, whereas Instant-Parameters performs only $28.00$.

%The aggregate values in Table~\ref{tab:numerical-results}, however, do not directly illustrate how this regret--refresh tradeoff develops as the synchronization threshold changes or how the resulting costs accumulate over time. 
\color{black}To further illustrate the regret--refresh tradeoff, we examine how cumulative regret and refresh cost vary with the synchronization threshold and evolve over time in Fig.~\ref{fig:action-value-results}. 
\color{black}
Fig.~\ref{fig:action-value-results} first illustrates the calibration
of \airsync under the true-action regret metric. Increasing the
parameter threshold $\Delta$ generally reduces the number of model
refreshes by allowing the cached model to tolerate larger accumulated
changes before synchronization. This reduction in refresh comes
at the cost of increased decision loss, producing the expected
regret--refresh tradeoff. Under this objective, the selected
\airsync threshold remains $\Delta=2$, while the \klairsync
threshold is adjusted from $0.2$ to $0.3152$.

Fig.~\ref{fig:action-value-results} shows that Instant-Param refreshes less frequently but incurs higher cumulative regret than \airsync and \klairsync, whose additional refreshes help limit accumulated decision loss.
Overall, by accumulating small model changes, our methods achieve lower total cost than Instant-Param.

%The regret trajectories in
%Fig.~\ref{fig:action-value-results} make the resulting performance
%difference visible. Although Instant-Param communicates substantially
%less frequently, its cumulative decision loss grows above those of
%\airsync and \klairsync. 
%\color{black}The additional refreshes help prevent decision loss that would otherwise accumulate under the non-accumulating baseline. %Overall, our methods achieve lower total cost than \emph{Instant-Parameters} policy.

%Overall, by accumulating small model changes, our methods achieve lower total cost than Instant-Param.

%our methods achieve lower true-action regret and total cost than Instant-Param, at the expense of more refreshes.\color{black}

%The additional refreshes performed by the cumulative policies therefore correspond to synchronization events that prevent part of the decision loss that would otherwise accumulate under the non-accumulating baseline.

%Overall, the experiments highlight the benefit of cumulative certificates. Param-AoIntSync and KL-AoIntSync accumulate consecutive small model changes, whereas Instant-Param reacts only to individual changes. This leads to more refreshes but lower true-action regret and total cost in the reported experiment.
\color{black}
\section{Conclusion}

We introduced AoInt, a decision-aware measure of model freshness,
and developed \klairsync and \airsync with
regret--refresh guarantees. Experiments show that both methods achieve
lower total cost than a
non-accumulating parameter-change trigger.  We note that \airsync avoids probe evaluation.

%We introduced AoInt, a decision-aware measure of model freshness, and developed \klairsync and \airsync with regret--refresh guarantees. Experiments show that both methods achieve lower observed mean true-action regret and total cost than a non-accumulating parameter-change trigger, at the expense of more refreshes. These results demonstrate the benefit of accumulating model changes when making synchronization decisions. We further note that \airsync avoids client-specific probe evaluation, providing a lightweight alternative to \klairsync.

%Future work can entail extending our framework to simultaneous model and
%environment drift, joint training and deployment, integration of application-centric objectives and trade-offs, and coordinated
%multi-device synchronization. Also, it can entail exploring continual federated learning checkpoints, certificate conservatism for
%larger models, adaptive distance measures and threshold calibration under changing model dynamics, and deployment in emerging settings such as spatial intelligence under smart AI glasses.

Future work includes simultaneous model and environment drift, joint training and deployment, application-centric tradeoffs, and coordinated multi-device synchronization. Further directions include continual federated learning checkpoints, certificate conservatism in larger models, adaptive distance measures and thresholds, and spatial intelligence with smart AI glasses.

%, at the expense of more
%refreshes. %Neither requires online labels or rewards, while \airsync
%avoids client-specific probe evaluation.

%We introduced AoInt, a decision-aware freshness measure, and developed \klairsync and \airsync with regret--refresh guarantees. Both reduce observed mean true-action regret and total cost over a non-accumulating trigger, at the cost of more refreshes, without online labels or rewards; \airsync further avoids client-specific probe evaluation.

%\vspace{-0.2cm}

\section{Acknowledgement}
This work was supported in part by the U.S. NSF grants 2106150, 2612077, 2549452, 2603236, ECCS-2543754, and by Meta Research. The authors acknowledge the University at Buffalo ORIED for GPU.
%\vspace{-0.3cm}

\bibliographystyle{ACM-Reference-Format}
\bibliography{references}

\end{document}